\documentclass[letterpaper, 10 pt, conference]{ieeeconf}
\IEEEoverridecommandlockouts
\usepackage{graphicx}
\usepackage{amsmath}
\usepackage{amssymb}
\usepackage[dvipsnames]{xcolor}
\usepackage{xspace}
\usepackage{url}
\usepackage{array}
\usepackage{booktabs}
\usepackage{siunitx}
\usepackage{makecell}
\usepackage{gensymb}

\usepackage{subcaption}
\usepackage{bytefield}
\usepackage{multirow}
\usepackage{glossaries}
\glsdisablehyper 
\usepackage[normalem]{ulem}
\usepackage{pgfplots}
\usepackage{tikz}

\usepackage{cite}       

\makeatletter
\let\NAT@parse\undefined
\makeatother

\usepackage[colorlinks]{hyperref}
\hypersetup{
    colorlinks = true,
    linkcolor  = MidnightBlue,
    citecolor  = RoyalPurple,
    urlcolor   = NavyBlue
}

\newcommand{\authorList}{Jason Hughes\textsuperscript{1},\xspace Marcel Hussing\textsuperscript{1}*, \xspace Edward Zhang\textsuperscript{1}*, \xspace Shenbagaraj Kannapiran\textsuperscript{1}, \\ Joshua Caswell\textsuperscript{1},
\xspace Kenneth Chaney\textsuperscript{1},\xspace Ruichen Deng\textsuperscript{1}, \xspace Michaela Feehery\textsuperscript{1}, \xspace Agelos Kratimenos\textsuperscript{1}, \xspace Yi Fan Li\textsuperscript{1}, \\\xspace Britny Major\textsuperscript{1}, \xspace Ethan Sanchez\textsuperscript{1}, \xspace Sumukh Shrote\textsuperscript{1}, \xspace Youkang Wang\textsuperscript{1},  \xspace Jeremy Wang\textsuperscript{1}, \xspace Daudi Zein\textsuperscript{1}, \\\xspace  Luying Zhang\textsuperscript{1}, \xspace Ruijun Zhang\textsuperscript{1}, \xspace Alex Zhou\textsuperscript{1}, \xspace Tenzi Zhouga\textsuperscript{1}, 
 \xspace Jeremy Cannon\textsuperscript{2}, \xspace Zaffer Qasim\textsuperscript{2}, \\\xspace Jay Yelon\textsuperscript{2}, \xspace Fernando Cladera\textsuperscript{1}, \xspace Kostas Daniilidis\textsuperscript{1}, \xspace Camillo J. Taylor\textsuperscript{1}, \xspace Eric Eaton\textsuperscript{1}}

\newif\ifdraftcolor

\draftcolortrue 

\ifdraftcolor

\else

\fi 

\newacronym{gnss}{GNSS}{Global navigation satellite system}
\newacronym{cots}{COTS}{commercial-off-the-shelf}
\newacronym{uav}{UAV}{unmanned aerial vehicle}
\newacronym{ugv}{UGV}{unmanned ground vehicle}
\newacronym{grd}{GRD}{ground resolution distance}
\newacronym{sfm}{SFM}{structure from motion}
\newacronym{hdr}{HDR}{high dynamic range}
\newacronym{fov}{FoV}{field of view}
\newacronym{agl}{AGL}{above ground level}
\newacronym{erc}{ERC}{event rate controller}
\newacronym{psnr}{PSNR}{peak signal-to-noise ratio}
\newacronym{ssim}{SSIM}{structural similarity index}
\newacronym{odm}{ODM}{OpenDroneMap}
\newacronym{utm}{UTM}{Universal Transverse Mercator} 
\ifdraftcolor

\else

\fi 

\title{A Multi-Robot Platform for Robotic Triage \\ Combining Onboard Sensing and Foundation Models}   

\author{\authorList
\thanks{*These authors contributed equally. Corresponding author: {\tt\small jasonah@seas.upenn.edu}.
\textsuperscript{1}Authors are with the GRASP Lab, School of Engineering and Applied Sciences, University of Pennsylvania. \textsuperscript{2}Authors are with the Perelman School of Medicine, University of Pennsylvania. 
This work was supported by the DARPA Triage Challenge under grant \#HR001123S0011. 
The views and conclusions contained herein are those of the authors and should not be interpreted as representing the official policies of DARPA.}}%
\begin{document}

\maketitle

\begin{abstract}
This report presents a heterogeneous robotic system designed for remote primary triage in mass-casualty incidents (MCIs). The system employs a coordinated air-ground team of unmanned aerial vehicles (UAVs) and unmanned ground vehicles (UGVs) to locate victims, assess their injuries, and prioritize medical assistance without risking the lives of first responders. The UAV identify and provide overhead views of casualties, while UGVs equipped with specialized sensors measure vital signs and detect and localize physical injuries. Unlike previous work that focused on exploration or limited medical evaluation, this system addresses the complete triage process: victim localization, vital sign measurement, injury severity classification, mental status assessment, and data consolidation for first responders. Developed as part of the DARPA Triage Challenge, this approach demonstrates how multi-robot systems can augment human capabilities in disaster response scenarios to maximize lives saved.
\end{abstract}

\section{INTRODUCTION}

Robotics has long sought to augment human capabilities in hazardous scenarios. Mass-casualty incidents (MCIs), such as those resulting from natural disasters, bombings, plane crashes, or industrial chemical spills, present an opportunity for robotic systems to assist first responders. The critical first step of providing medical assistance during MCIs is {\em primary triage}: the initial process of locating victims at the site of the MCI and assessing the severity of their injuries to prioritize treatment, which is essential to optimizing survival outcomes. Traditionally, primary triage relies on human responders who may face significant risk and information overload \cite{bergen2020stress}, underscoring the potential for automated systems to mitigate these challenges.

While prior efforts have explored the use of air-ground robotic teams for search and exploration in disaster zones \cite{cladera2024challenges, miller2022stronger, luo2011airground, hsieh2007adaptive}, few systems have focused specifically on rapid triage. Existing approaches typically solve parts of the problem in isolation without integrating comprehensive triage functions. For example, air-ground teams have also been developed to find and localize objects of interest \cite{miller2022stronger, miller2020minetunnel} or localizing injured individuals without triaging the victims \cite{song2023design}. Toward autonomous triage, \cite{senthilkumaran2024artemisaidrivenrobotictriage} employ a UGV to assess injuries, but do not address the task of locating victims or measuring crucial vitals. Similarly, \cite{gutierrez2024design} developed a low-cost robot that estimates patient height, weight, heart rate, and temperature, but does not assess physical injury severity. Furthermore, recent initiatives such as the DARPA Triage Challenge (DTC)~\cite{DTCWebsite} emphasize the need for fully remote systems capable of operating in dynamic MCI environments to identify victims, assess vital signs, alertness, and injuries, and prioritize treatment based on severity. 

\begin{figure}[t!]
    \centering
    \includegraphics[width=\linewidth]{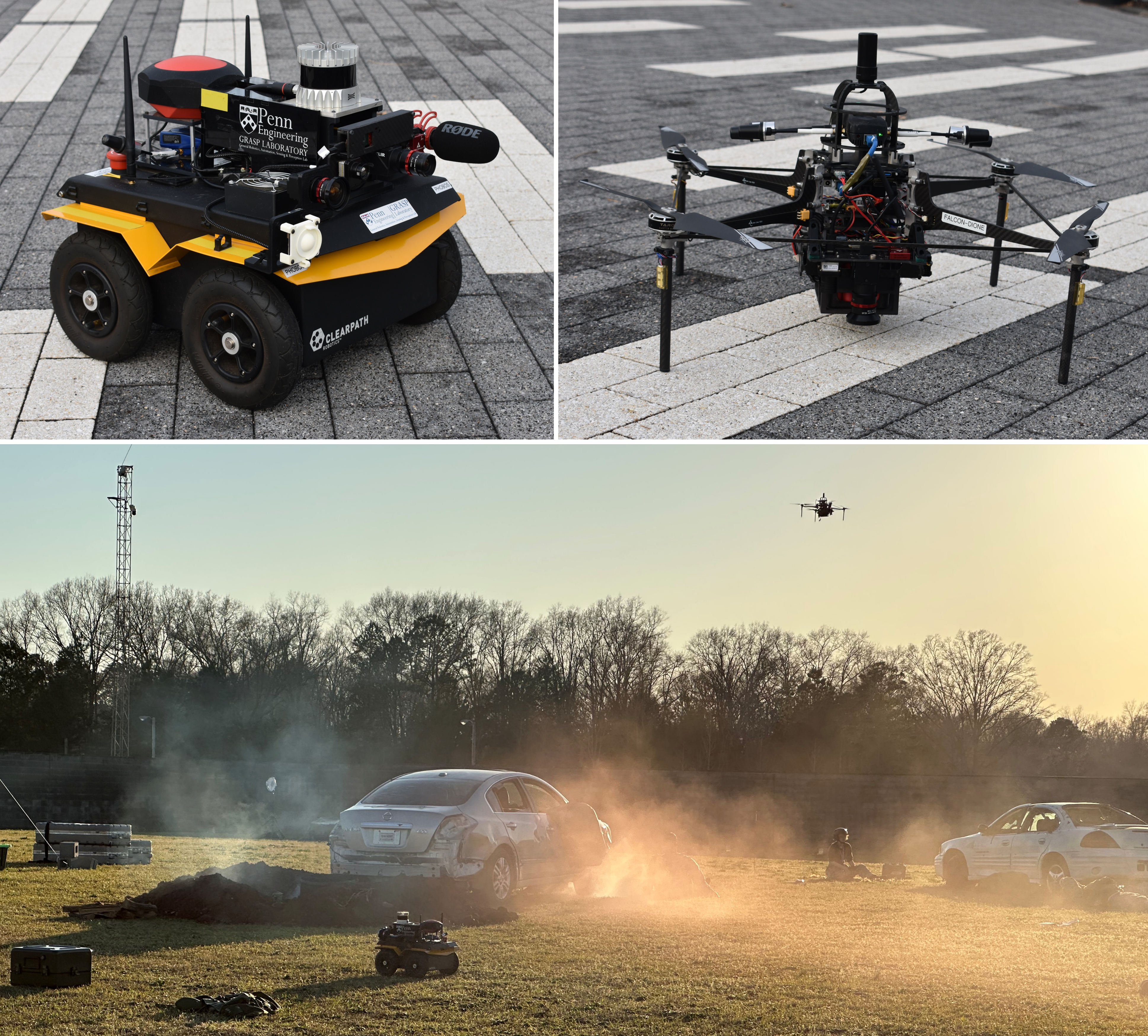}
  \caption{Our heterogeneous robot team for triage during a mass casualty simulation at the Guardian Centers, Perry, GA in March 2025.}
  \label{fig:platforms}
\end{figure}

In this technical report, we present a heterogeneous air-ground team toward remote primary triage in MCIs developed as a part of the DARPA Triage Challenge, specifically focusing on the system developed for the first and second years of the competition with the main differences between the first and second years being the scale and difficulty of the conditions. The second year of the competition required teams to operate on larger incidents, both in terms of area and number of victims, and required teams to operate at night. The challenge requires teams to find, localize, and triage victims in mock MCIs. For triaging, teams are required to assess victims' vitals (e.g., heart rate, respiration rate), alertness (e.g., ocular, motor, verbal), and physical injuries. In the challenge, teams submit their assessments to DARPA for scoring but this assessment could also be sent to first responders.

Our system uses unmanned aerial vehicles (UAVs) to quickly detect and localize people, and obtain a top-down view of each person. The unmanned ground vehicles (UGVs) then navigate to each person. Providing a stationary platform for stable sensing, the UGVs use a suite of cameras and sensors to assess each victim. Sensor data is analyzed using a combination of on-board unimodal algorithms and multi-modal learned methods. We designed a system that allows us to develop algorithms in a plug-and-play fashion to quickly iterate on them. We have two main categories of algorithms. First, we have algorithms that process time-series data for heart rate detection and respiration rate detection. Second, we leverage the rise of foundational models as general-purpose machine learning tools~\cite{vaswani2017attention, kojima2022large, brown2020language}. We fine-tune and test open-source vision-language models (VLM) of various sizes, to perform injury identification based on multi-view image and audio data. The VLM also provides the added benefit of describing each victim to assist first responders in locating them.  All of this derived information is synthesized into a comprehensive assessment of each victim’s status that can then be reported to first responders. This report highlights our multi-robot system designed to facilitate rapid primary triage via a highly cooperative and interconnected heterogeneous robot team. 

\section{SYSTEM OVERVIEW}
This section outlines the hardware and software components deployed across our platforms and emphasizes the modular architecture that supports fast integration and flexible reconfiguration of the system.

\subsection{Hardware}
\begin{figure}[htbp]
    \centering
    \begin{subfigure}{0.46\textwidth}
        \centering
        \includegraphics[width=\textwidth]{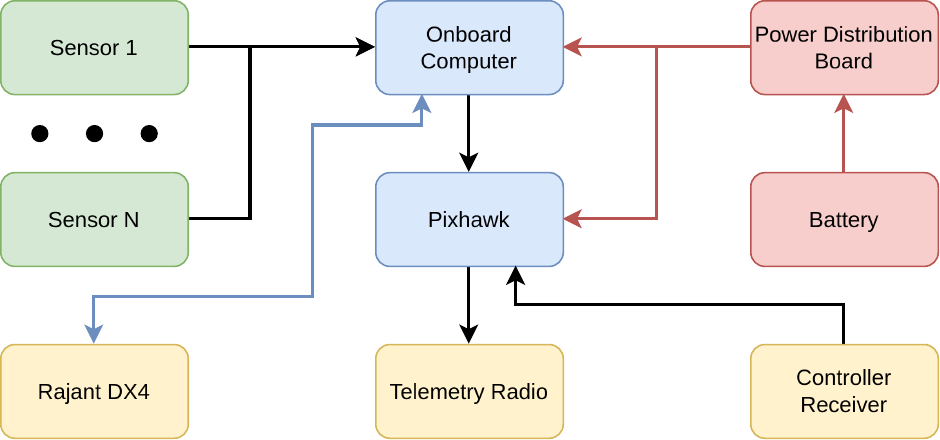}
        \caption{UAV Hardware Diagram}
        \label{fig:uav_hardware}
    \end{subfigure}
    \hfill
    \begin{subfigure}{0.46\textwidth}
        \centering
        \includegraphics[width=\textwidth]{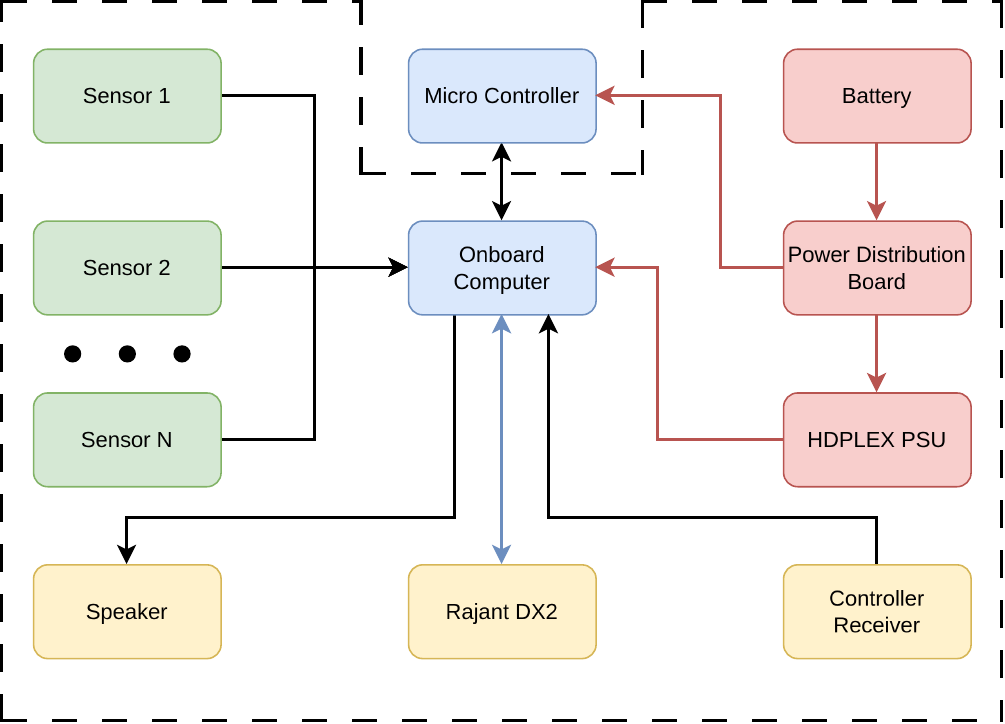}
        \caption{UGV Hardware Diagram}
        \label{fig:ugv_hardware}
    \end{subfigure}
    \caption{These diagrams outline the hardware setup for both the UAV (a) and UGV (b). The sensors (green) connect to the onboard computer via USB. Each robot is equipped with a Rajant radio connected via ethernet (blue arrow), and other communication hardware (yellow) is there for control and safety. Finally, power (red) is provided from a battery to the onboard computer via a power distribution board. The dotted outline on the UGV indicates modularity, everything within the box can be taken and placed on another robot and only one connection to that new robot is needed.}
    \label{fig:hardware}
\end{figure}
Our platform consists of an air-ground team using the Falcon 4 UAV, built in-house, several Clearpath Jackal UGVs and a basestation. A diagram showing the specifics of the robot hardware configurations is provided in Fig. \ref{fig:hardware}.

As our eye-in-the-sky, we use a custom UAV platform called the Falcon 4. It is a 673.3\,mm quadcopter designed for high-altitude overhead flight, more information on the platform can be found in \cite{liu2022denseforest}. We outfitted the quadcopter with an NVIDIA Jetson Orin NX with 16\,GB of unified memory. We use a 3.2MP FLIR Blackfly-S RGB camera with a 5\,mm lens for daytime imaging and a FLIR Boson+ longwave infrared (LWIR) camera for night time imaging. The UAV is also outfitted with a Rajant DX4 mesh radio for communications and a u-blox ZED-F9P GPS and  VectorNav VN100T-CR IMU for localization. All the low-level control for the UAV is handled by the ARC V6X, a Pixhawk-based flight controller.

On the ground, we employ multiple Clearpath Jackal UGVs, which are outfitted with an AMD Ryzen 7 8700G CPU with 64\,GB of RAM and an NVIDIA RTX 4000 SFF Ada GPU with 20\,GB of VRAM. The computer is powered by the HDPLEX 500\,W Hi-Fi DC-ATX, a DC-to-DC power supply. We tested a number of sensors that cover most of the safe electromagnetic spectrum. Not all of these sensors were implemented on the final versions of the UGV systems that were deployed in competition. For camera sensors, we used the same FLIR Blackfly-S 3.2MP color camera and Boson+ LWIR camera as the UAV, although the RGB camera has an 8\,mm lens on the UGV. We also tested a Prophesee EVK4 event camera for respiration rate detection. For radar, we used an Acconeer XE125 for respiration rate detection and later upgraded to a Texas Instruments IWR6843 millimeter wave radar for both respiration and heart rate detection. We use the u-blox ZED-F9P GPS for global localization and a VectorNav VN-100T-CR IMU for orientation. We also use a speaker to play instructions to the patient and use a shotgun microphone to record their responses. We use a Rajant DX2 mesh radio to talk between the UAV and the basestation. Finally, we have an Ouster OS1-64 for future autonomy capabilities. Any subset of these sensors is responsible for triaging when the system is deployed.



\subsection{Software}
\begin{figure}[t!]
    \centering
    \begin{subfigure}{0.45\textwidth}
        \centering
        \includegraphics[width=\textwidth]{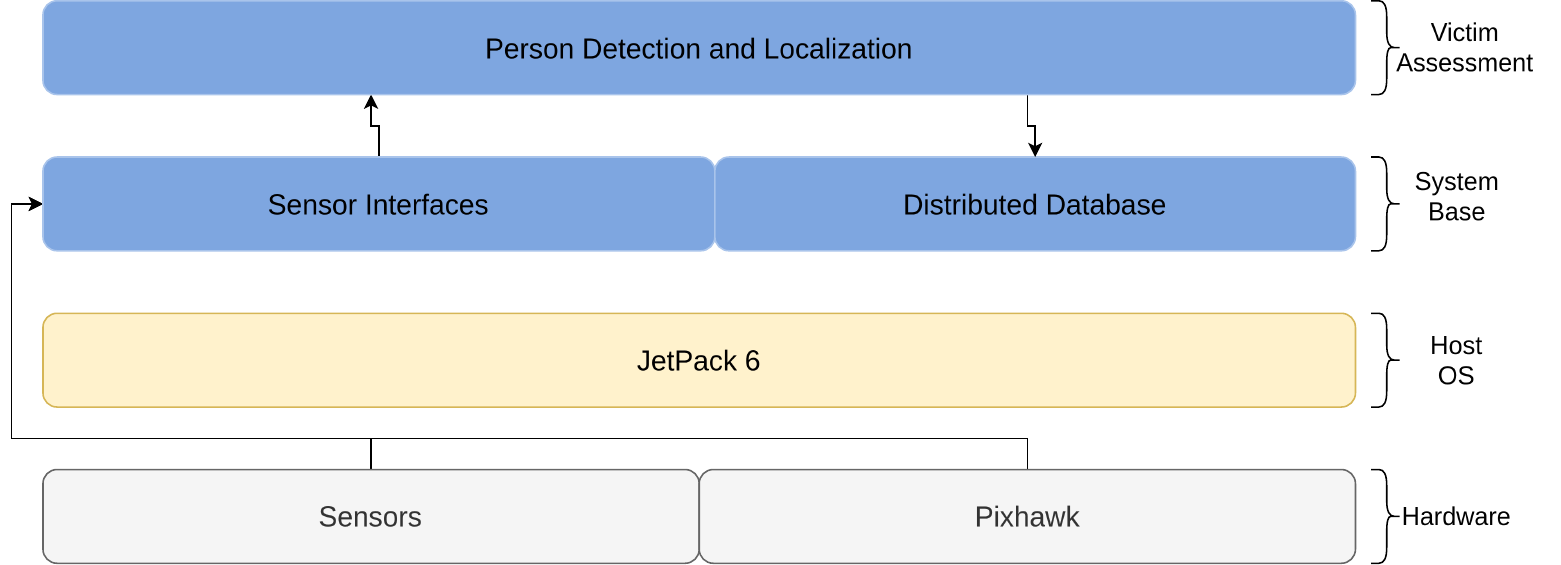}
        \caption{UAV Software Diagram}
        \label{fig:uav_software}
    \end{subfigure}
    \vspace{5mm}
    \begin{subfigure}{0.45\textwidth}
        \centering
        \includegraphics[width=\textwidth]{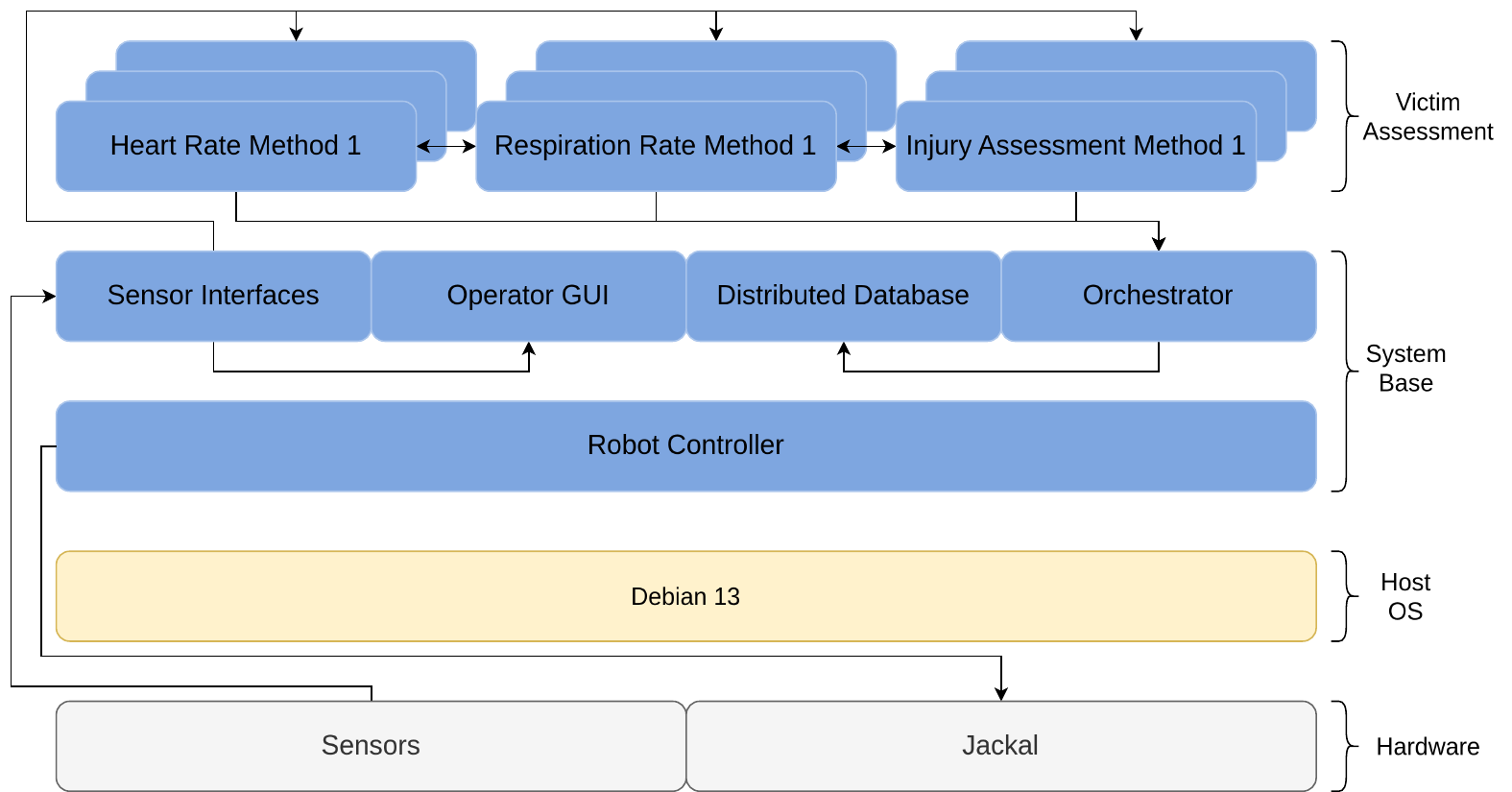}
        \caption{UGV Software Diagram}
        \label{fig:ugv_software}
    \end{subfigure}
    \vspace{5mm}
    \begin{subfigure}{0.45\textwidth}
        \centering
        \includegraphics[width=\textwidth]{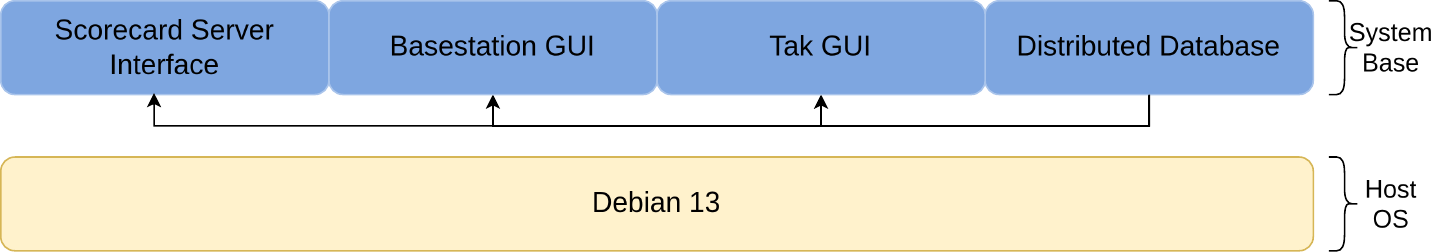}
        \caption{Basestation Software Diagram}
        \label{fig:basestation_software}
    \end{subfigure}
    \caption{This depicts how the software is laid out on each platform. The hardware level consists of
    the sensors, the onboard computer and the robot itself. All of our software runs on the host OS (yellow) of the onboard computer. All software is modularized into docker images (blue), which we depict in two levels on the UAV and UGV. The system base level takes care of hardware interfaces and data management. The top layer is for victim assessment. The arrows depict the flow of data starting at the sensor hardware.}
    \label{fig:software}
\end{figure}



Our software design revolves around modularity. Our objective with this design was to easily add and drop algorithms necessary to triage from new sensors. We achieved this modularity using Docker and Robot Operating System (ROS)~\cite{Quigley09ros}. We can develop the algorithm in its own container with its own dependencies and then deploy this container on the robot with docker compose. All the docker images are designed to look at a boolean environment variable that we can use to toggle the container on and off easily. When the container is active it can receive sensor data and publish results with ROS topics. We then have an orchestrator that looks at what docker images are running, and waits for the results of each algorithm and aggregates all the outputs into what we call a scorecard. This allows us to easily test and deploy new sensors on the robot and easily replicate setups on new robots. 

Another core design feature is data transmission reliability. Our system design allows us to operate in scenarios with degraded communications. We use the UAV to spot people and provide a prior as to where each UGV should go, which is done solely to optimize the triage time and is not strictly necessary. We also need to reliably transmit triage outcomes from the robot back to a basestation or first responders. To this end, we use a distributed database, MOCHA \cite{cladera2024mocha} to transmit data opportunistically. When a robot finds a victim or completes triaging them, it stores the results in its database. When the Rajant mesh radios have a strong connection to one another the databases are synced. This ensures we only send data when communication is reliable. It also allows data to hop along the robots to get back to a basestation or first responder. 

We need our triaging to be robust to the scenarios of the challenge and real-life triage deployments, which includes being able to operate at night without external illumination. Since many triage components rely on RGB imagery we leverage the large Sony Starvis image sensor in the Blackfly-S to maximize low-light performance. We follow a traditional high-dynamic-range (HDR) pipeline: the system first captures a single image with 10 millisecond exposure to estimate scene illumination, then automatically adjusts exposure time, gain, and gamma via a closed-loop metering procedure that is subject to camera limits. Once the target brightness is reached, the camera captures a short burst of frames at multiple exposure levels -- bright (long exposure), medium, and dark (short exposure), which are fused into an HDR composite that preserves details in both shadows and highlights. This fused HDR image is then published on a ROS topic and used in place of the raw RGB frames by the downstream perception modules, including the injury assessment and ground-level casualty localization, enabling reliable operation in low-light conditions.

Triaging is broken down into two main parts, heart rate and respiration rate estimation, and injury localization and classification. The software architecture for all the heart rate and respiration rate algorithms works similarly. The data is collected from the sensor using its SDK, we then use a ROS driver to publish the sensor data on a ROS topic. When a trigger is issued by the user, indicating that the UGV is positioned correctly relative to the person, the algorithm subscribes to the sensor data topic to collect the time-series data for ten to fifteen seconds, depending on the algorithm. This data is then fed through a neural network or it is augmented and traditional signal processing algorithms are applied to it. We detail each of the methods we use for heart rate and respiration rate in Sec.~\ref{sec:va}. For injury localization and classification we use individual images and large foundation models such as fine-tuned VLMs and DINO-based classifiers, which are detailed in Sec.~\ref{sec:vlm}.

\subsection{User Interfaces}
The system provides a few interfaces that assist in monitoring its individual components during operation.
\subsubsection{Team Awareness Kit}

To support situational awareness during triage operations, specifically geared towards first responders, we provide a map-based interface that presents real-time information about robot position, detected casualties and their triage assessments. The interface is built on the Android Team Awareness Kit (ATAK)~\cite{usbeck2015improving}, a widely deployed platform used by first responders and military operators. ATAK offers a map-centric operational picture, communication tools, and a plugin architecture that enables custom extensions on Android devices. Robot positions and triage assessments are collected by the basestation via MOCHA, we then use ROSTAK~\cite{rostak} to reformat the ROS topics into the Cursor-on-Target (CoT) protocol. A TAK server runs on the basestation that all edge devices connect to, specifically the Android devices with our custom plugin that first responders would carry.
We extend ATAK’s native functionality to better support mass-casualty triage. Custom iconography is used to distinguish robots from casualties, and the color of the casualty icon encodes the triage severity. A dedicated triage panel summarizes all detected casualties and selecting a casualty marker opens a detailed report that includes the full triage report and an image of the casualty. 


\subsubsection{Operator Overview}

We provide a browser-based interface that can be used to remotely monitor the UGV's triage status and enables operators to view RGB and IR video feeds with low latency. A FastAPI~\cite{FastAPI} server is hosted by each UGV. Using WebRTC for transport, the system encodes incoming images from a ROS topic with H.264 hardware-accelerated codecs to ensure smooth, real-time viewing. A user can send configuration commands over a TCP interface to allow for browser-side control of camera mode, frame rate, resolution, and image filtering, such as CLAHE contrast enhancements for IR imagery. 

In addition to video, the webpage incorporates additional situational-awareness tools. A map viewer panel displays the robot position and heading. There is also a status window showing the status of the triage in process. These views of first-person, global context, and triage decisions support safer and more informed operation of each robot. 

\subsubsection{System Monitoring}

\begin{figure}
    \centering
    \includegraphics[width=\linewidth]{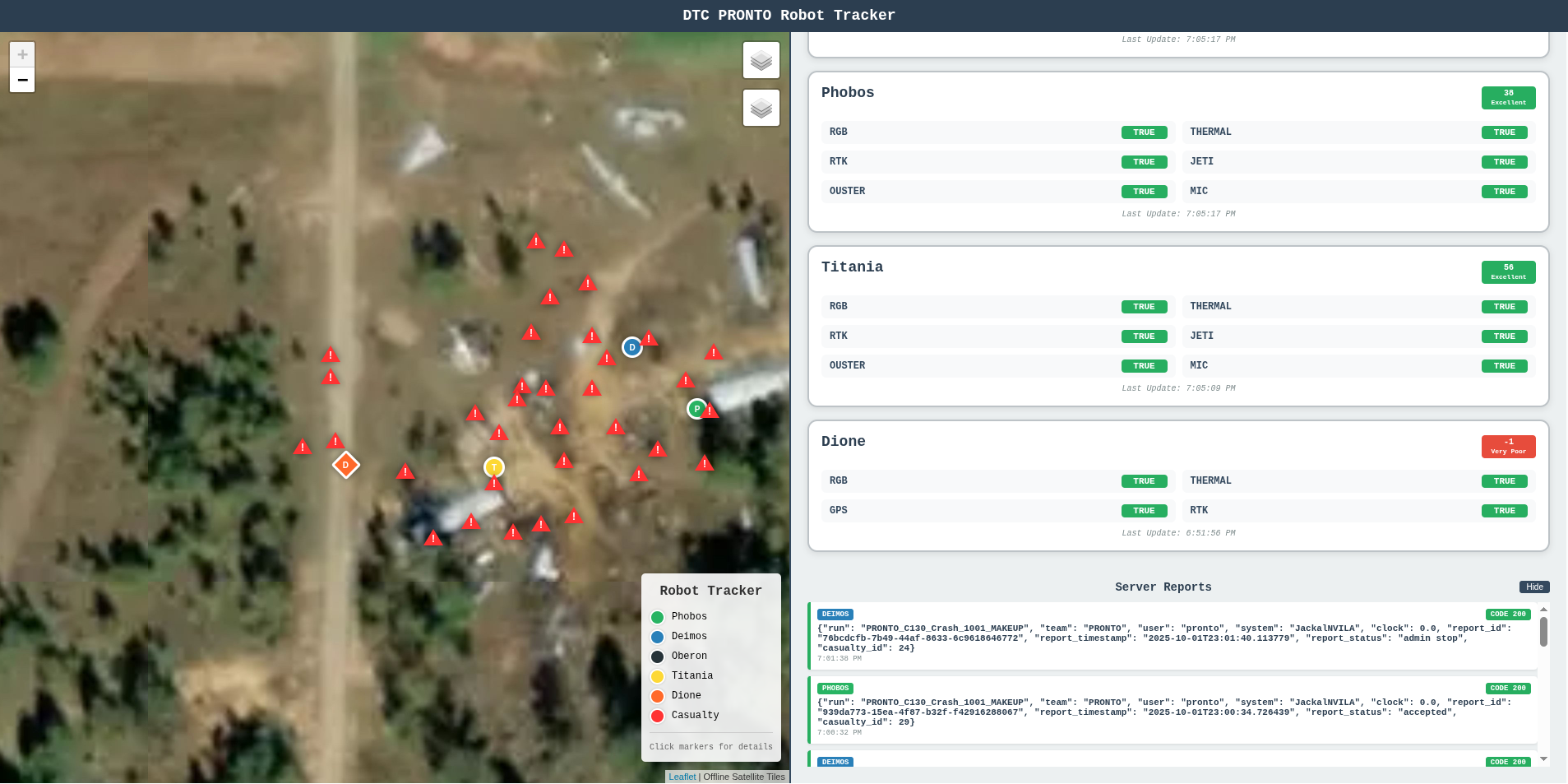}
    \caption{System Monitoring Interface from the DARPA Triage Challenge, held at Guardian Centers, Perry, GA in October 2025.}
    \label{fig:geoviz}
\end{figure}

Finally, we develop a GUI for monitoring the system as a whole. The basestation runs an instance of the distributed database, allowing it to get asynchronous updates from all the robots. The robots share their position, the status of their sensor stack and triage information from victims. The robot and victim positions are plotted on satellite imagery. The robot sensor health is displayed for each robot in the system. Finally, when in competition we display http status codes from the DARPA scoring server to ensure we are communicating with it. An example of the interface from Challenge Event 2 of DTC is shown in Fig.~\ref{fig:geoviz}.

\section{VICTIM LOCALIZATION}
\label{sec:vl}
In this section we discuss how we localize victims on both the UAV and UGV.

\subsection{Aerial Localization}
To rapidly provide victim assessments across large MCIs, the UAV performs continuous overhead detection and geolocalization of victims in both daytime and nighttime conditions. To enable accurate detection of victims on the ground we place both the LWIR and RGB cameras on a dampened plate on the front of the UAV with its own VectorNav VN-100T-CR IMU. To first detect victims, we fine-tuned two YOLO-based~\cite{redmon2015unified} detectors, one for RGB imagery during the day and one for LWIR imagery at night. For nighttime detection of victims we started with a YOLOv8~\cite{yaseen2024yolov8} model which we tuned on the datasets in \cite{jia2021llvip, flir-adas_dataset, utokyo-multispectral-dataset} and data we collected ourselves from flying the platform. Since IR imagery is 14-bit we first normalize it down to 8-bit and then apply a CLAHE filter to the image before feeding it to the network both at train time and test time. For daytime detection we start with a pretrained YOLOv8 model and train only on data we collect from our UAV.  


Once a victim is detected, we project the 2D bounding box center into the world frame using the UAV's GPS, IMU orientation, camera intrinsics and camera-IMU extrinsics. Each pixel is back-projected to a ray in the camera frame, transformed into the world frame using the current camera orientation, and intersected with the ground plane to obtain a 3D position estimate. This pixel-to-world conversion runs online for every frame, enabling continuous updates to victim locations even as the UAV changes position.
Since the UAV revisits targets multiple times during flight, detections are temporally fused via spatial clustering, and a persistent casualty ID is assigned to each casualty. Each new detection is associated with an existing cluster if it falls within 2\,m of a cluster center; otherwise a new cluster (and ID) is created. Cluster centers are updated using a weighted average that places higher confidence on detections appearing near the center of the image, where reprojection error is lower. The resulting aerial victim map containing validated cluster locations, their corresponding casualty IDs, and cropped victim imagery is synchronized through MOCHA and shared with UGVs to guide navigation and downstream triage operations.

\subsection{Ground Localization}
On the ground, the UGV refines casualty locations and establishes a consistent mapping between aerial detections and the individuals encountered during triage.

We first run a YOLOv8-based instance segmentation model on the UGV's RGB image to isolate the casualty. At night, we use the HDR image from the camera. From the predicted mask, we compute the 2D centroid in pixel coordinates. Using the calibrated camera intrinsics, the pixel is converted into a horizontal bearing relative to the camera's optical axis. LiDAR points whose azimuth falls within a narrow angular window around this bearing are then selected, effectively cropping the LiDAR scan to the region containing the casualty. Within this angular slice, we cluster the LiDAR points and take the centroid of the nearest cluster as the casualty position in the UGV’s local frame. This local displacement is transformed into the world frame using the UGV’s IMU heading and GPS position to obtain a global casualty estimate.

To associate this ground-level estimate with an existing aerial detection, we perform nearest-neighbor matching in the global frame: if the estimate lies within 2\,m of a UAV-derived casualty location, we assign the same casualty ID; otherwise, we instantiate a new casualty ID.


\section{VITALS ASSESSMENT}
\label{sec:va}

A key aspect of primary triage is assessing the victim's vital signs, we focus on heart rate and respiration rate. In this section, we discuss the different sensors and associated algorithms we used to remotely detect these vital signs.




\subsection{Heart Rate Detection}\label{sec:heart-rate}
We tested three methods to remotely detect heart rate: remote photoplethysmography (rPPG), a pre-trained neural network and a millimeter wave (mmWave) radar. 

\subsubsection{rPPG}

Remote photoplethysmography (rPPG) is a contactless technique that estimates heart rate by analyzing the periodic RGB color changes in images caused by blood volume fluctuations in human skin. While there are off the shelf rPPG packages, they are not designed to be deployed outside of clinical settings. We introduce two critical enhancements to adapt it for real-world triage scenarios, where motion~\cite{li2022motion}, lighting conditions~\cite{shao2025lighting} and occlusions~\cite{nguyen2023non,nguyen2024eval} are common challenges for rPPG. 

\begin{figure}[t!]
    \centering
    \begin{subfigure}[b]{0.20\linewidth}
        \includegraphics[width=\linewidth]{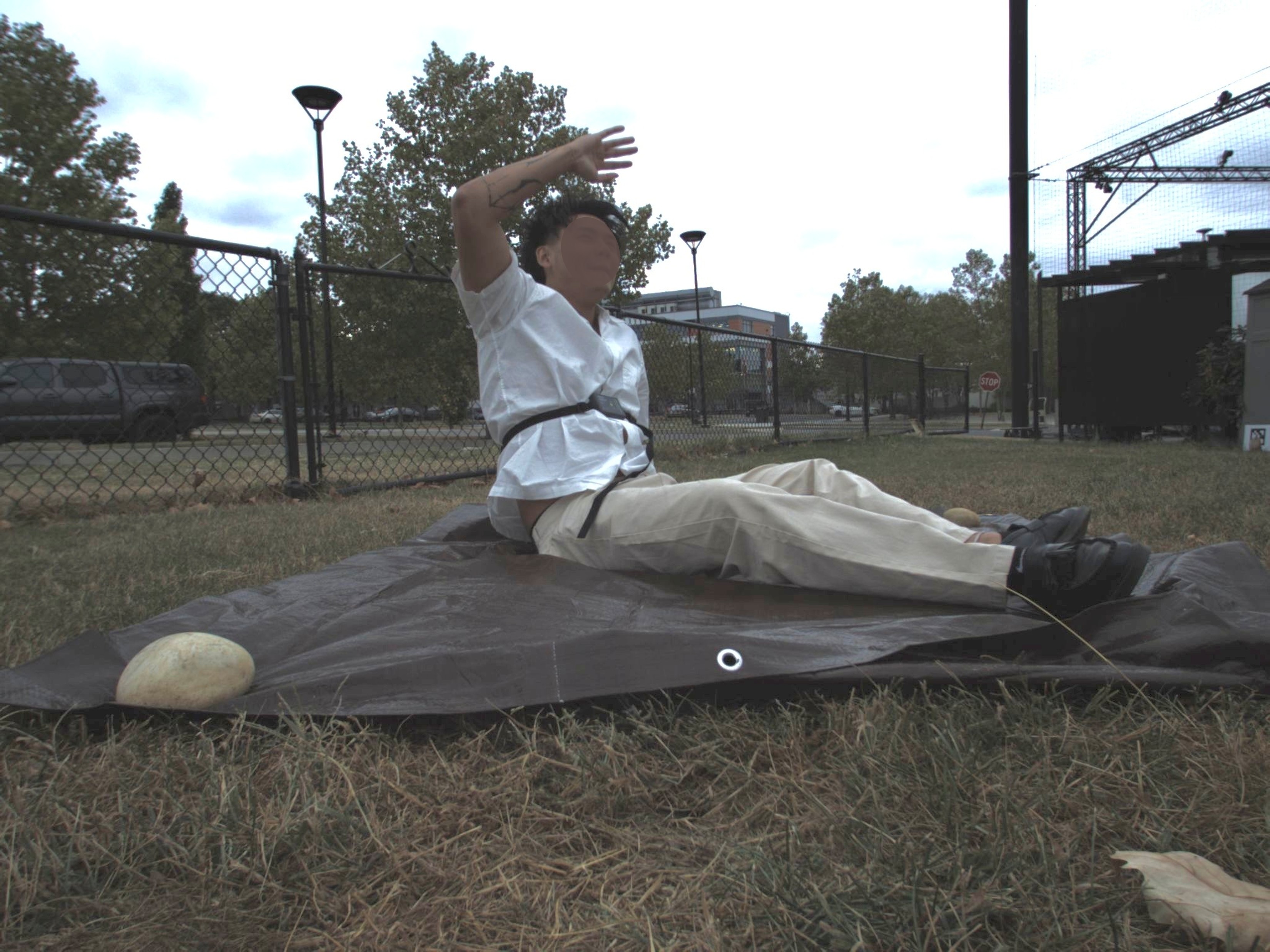}
        \caption{}
        \label{fig:original}
    \end{subfigure}
    \hfill
    \begin{subfigure}[b]{0.25\linewidth}
        \includegraphics[width=\linewidth]{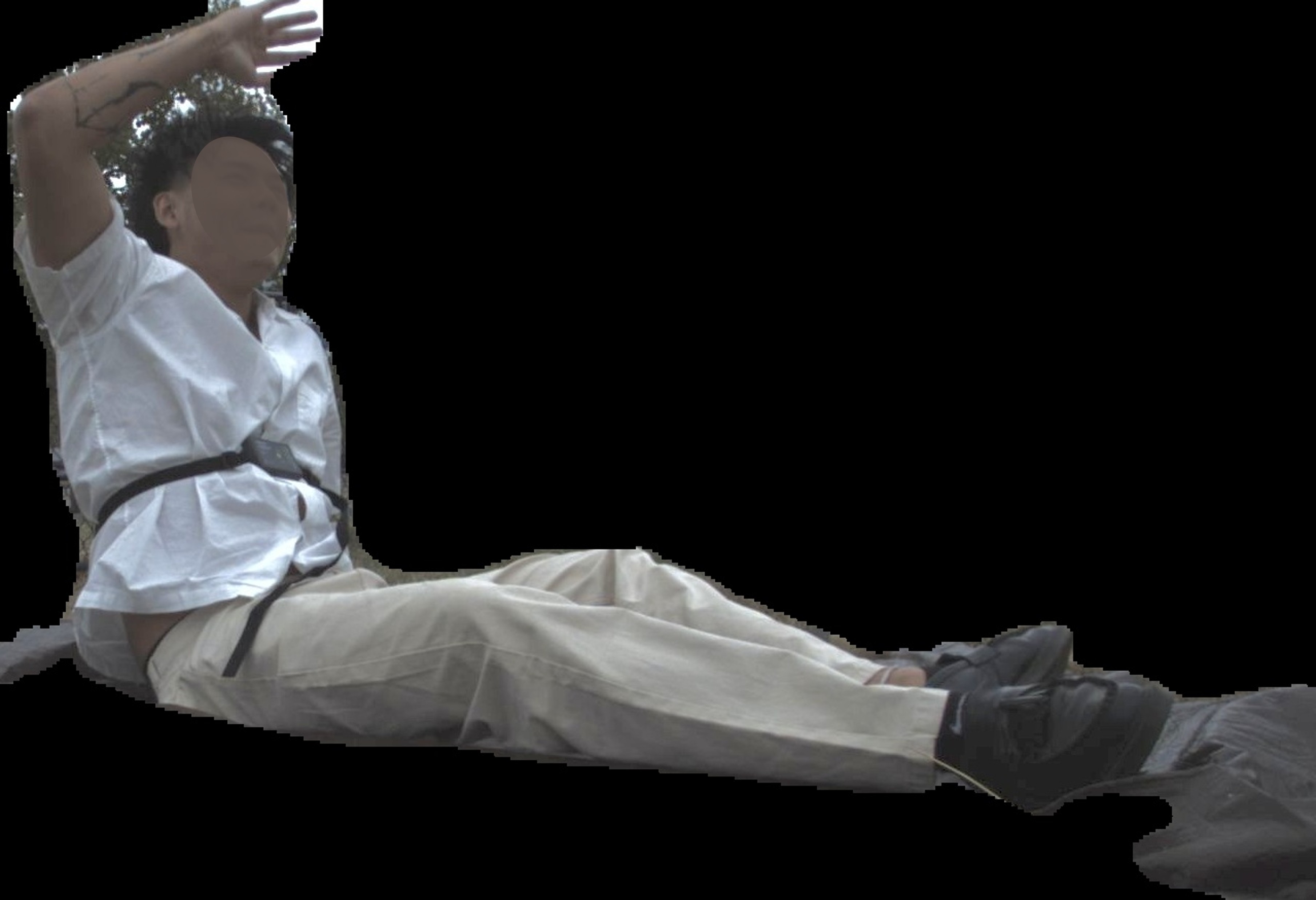}
        \caption{}
        \label{fig:cropped}
    \end{subfigure}
    \hfill
    \begin{subfigure}[b]{0.25\linewidth}
        \includegraphics[width=\linewidth]{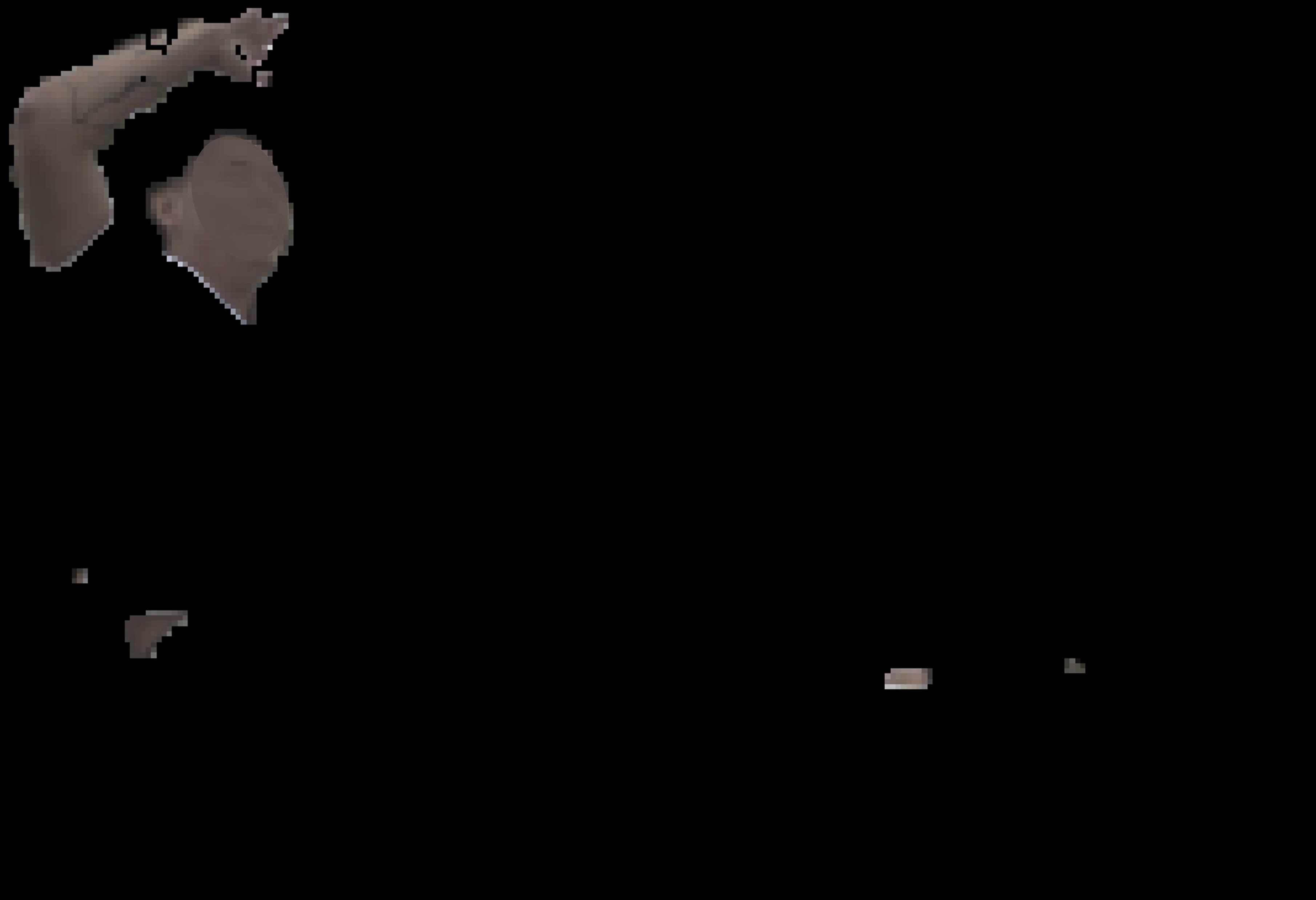}
        \caption{}
        \label{fig:skin}
    \end{subfigure}
    \hfill
    \begin{subfigure}[b]{0.25\linewidth}
        \includegraphics[width=\linewidth]{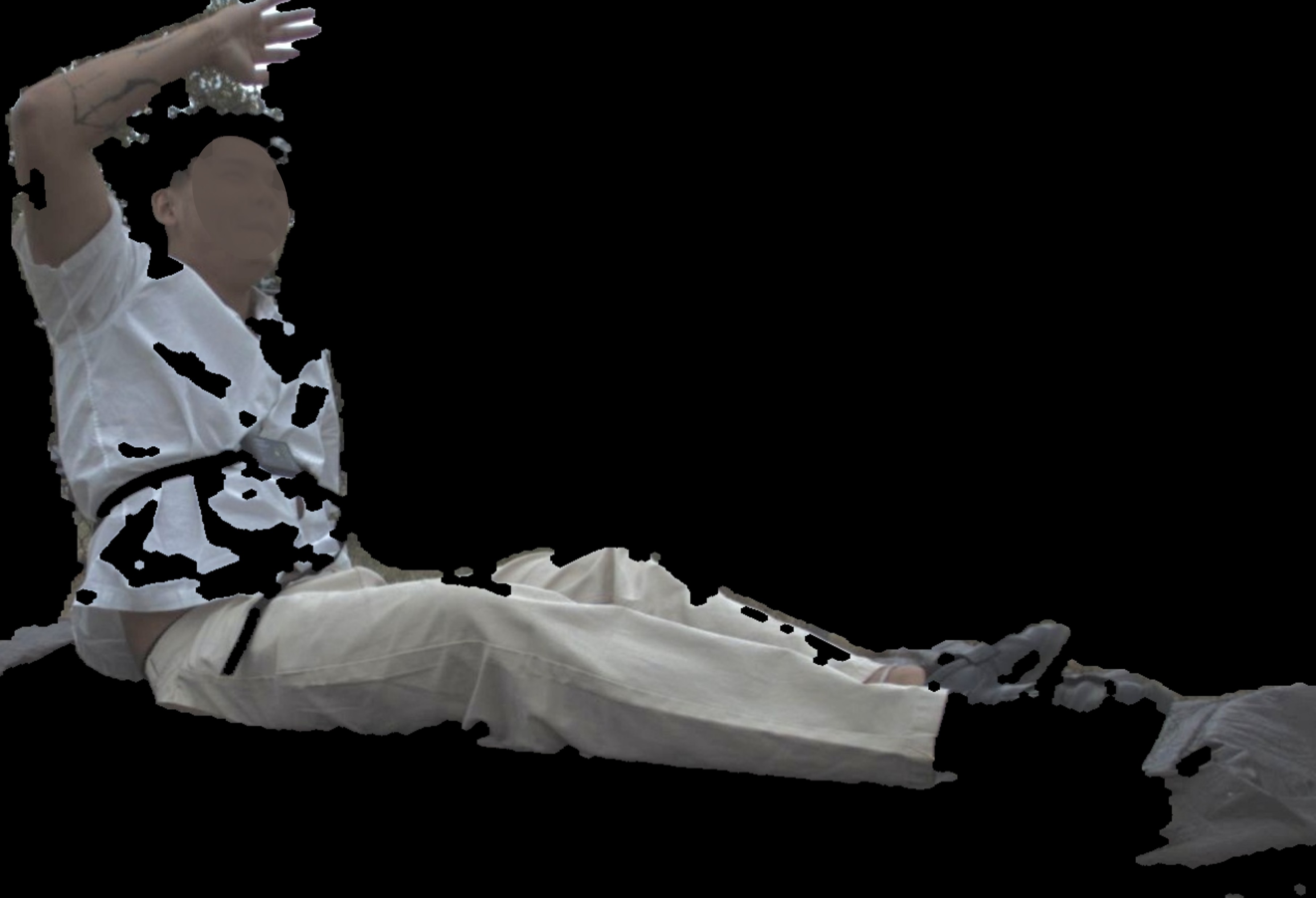}
        \caption{}
        \label{fig:rgbskin}
    \end{subfigure}
    \caption{Preprocessing image using YOLOv8, HSV Skin Extraction Results and RGB Skin Extraction. (\subref{fig:original}) Original image of a person sitting on the grass. (\subref{fig:cropped}) Cropped image using YOLOv8. (\subref{fig:skin}) Extracted skin image in HSV color space. (\subref{fig:rgbskin}) Extracted skin image in RGB color space, illustrating the difficulty of skin isolation in RGB. 
    }
    \label{fig:yolov8}
\end{figure}

First of all, field-acquired videos in triage environments often contain background clutter that introduces significant noise to rPPG signals. To mitigate this, our pipeline employs YOLOv8~\cite{yaseen2024yolov8} to isolate the victim from environmental interferences, as shown in Fig.~\ref{fig:yolov8}\subref{fig:cropped}. 


While conventional rPPG methods predominantly rely on facial regions of interest (ROIs), we adopt a more robust approach informed by the findings in occluded healthcare scenarios~\cite{wang2024chaos}. Therefore, we developed a non-facial skin-region selection method based on HSV (Hue, Saturation, Value) color-space analysis, which provides distinct advantages over traditional RGB processing shown in Fig.~\ref{fig:yolov8}\subref{fig:skin} and Fig.~\ref{fig:yolov8}\subref{fig:rgbskin}. The cylindrical-coordinate system of HSV enables a clear definition of skin tone boundaries through a continuous range. 

We apply this pipeline to 10 seconds worth of frames, totaling about 300 frames, and then extract the green channel from each of the frames due to its higher hemoglobin absorption sensitivity. Among many common methods for remote PPG measurements~\cite{boccignone2022pyvhr}, we choose a Chrominance-based method (CHROM)~\cite{Haan2013chrom} for the extraction of blood volume pulse (BVP) signals, which is crucial to the estimation of heart rate. Finally, we obtain the heart rate estimation in beats per minute (BPM) through post-filtering, where the BVP signals are passed through a Band-Pass filter in order to discard the frequencies that are beyond the range of typical heart rates.


\subsubsection{MTTS-CAN}
There have been many learning based approaches to heart rate detection, including \cite{liu2023rppgtoolboxdeepremoteppg, chen2018deepphysvideobasedphysiologicalmeasurement, 9010965, 9879451, 10030453, liu2021multitasktemporalshiftattention}. Among these MTTS-CAN~\cite{liu2021multitasktemporalshiftattention} emerges as a top performer, achieving the lowest mean absolute error and mean percentage error in heart rate estimation on public datasets~\cite{liu2021multitasktemporalshiftattention}. We use the model as is. At runtime, we collect 10 seconds worth of images at 30 Hz, totaling 300 images. The images are then fed to the network, which produces a signal for the 10 second interval. We run a peak finding algorithm on the signal to get the heart rate for the 10 second duration which we then convert to beats per minute.


\subsubsection{mmWave Radar}

We employ a Texas Instruments IWR6843AOP millimeter-wave (mmWave) radar for contactless heart-rate estimation, leveraging the on-chip vital-signs firmware that comes with the device. Similar Frequency Modulated Continuous Wave (FMCW) radars have been widely used for non-contact cardio–respiratory monitoring \cite{du2022vital, ahmad2018vital, Peake2023VitalSigns, wang2020remote}. The radar transmits FMCW chirps, and the firmware selects a static range bin at the subject’s chest to produce a displacement time series along the line of sight. This signal is band-pass filtered around the expected heart-rate band (1–3\,Hz), and a spectral peak detector yields a heart-rate estimate in beats per minute. A ROS driver publishes these outputs, and the downstream orchestrator rejects low-quality frames and maintains a sliding window over recent valid samples. 

\subsection{Respiration Rate}\label{sec:resp-rate}

In addition to heart rate, we also aim to detect respiration rate. We tested five methods to do so: a method based on pulsed coherent radar (PCR), the same deep network, MTTS-CAN chosen for heart-rate detection, the same mmWave Radar, using the LWIR camera, and using an event camera 

\subsubsection{Pulsed Coherent Radar}
We use an Acconeer A121 pulsed coherent radar (PCR) sensor to measure respiration rate. The radar emits a signal and measures its return, or relative permittivity, which differs based on the material the signal reflects from; by looking at this value, we can determine if the radar is pointed at a person (clothing or skin)~\cite{AcconeerDocs}. We can also use the return time to measure distance to the person. Once we have the distance to the person, we can measure micro-fluctuations in this distance over time, which includes movement in the chest due to breathing. This signal is passed through a bandpass filter to reduce noise and then we can do peak-finding to acquire a respiration rate from the signal.

\subsubsection{MTTS-CAN}
The MTTS-CAN model is also capable of producing a respiration rate signal from RGB imagery. We collected 10 seconds worth of images at a 30\,Hz frame rate, totaling 300 images. The images are then fed to the network. We then employ a standard peak finding algorithm to get the respiration rate for that 10 second duration. We then convert this to breaths per minute.

\subsubsection{mmWave Radar}

The same mmWave processing pipeline from heart-rate estimation also provides respiration-rate estimates from the lower-frequency component of the chest-motion signal. Respiration induces larger, slower oscillations (typically $0.1$–$0.5$\,Hz) that are well suited to mmWave displacement sensing. The firmware applies a low-frequency band-pass filter and spectral analysis to this signal and reports the dominant breathing rate in breaths per minute. We reuse the same ROS interface and gating logic for heart rate and again take the median over a sliding window of valid measurements. This modality remains reliable when vision-based respiration methods degrade due to low light, occlusion, or limited visible skin, and thus forms an important component of our overall vitals-estimation stack.

\subsubsection{LWIR Camera}

\begin{figure}
    \centering
    \includegraphics[width=0.98\linewidth]{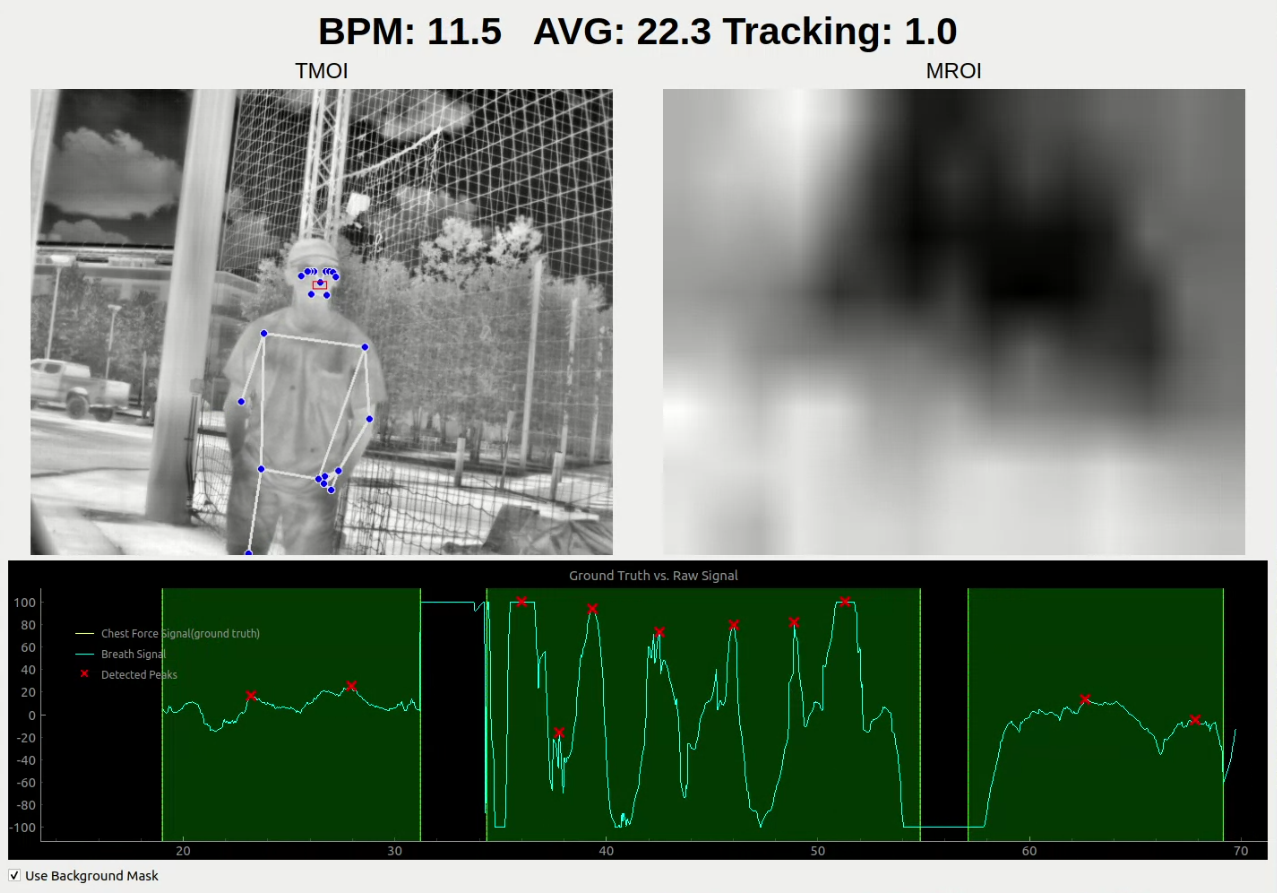}
    \caption{Results of our thermal respiration rate tracker in an outdoor environment demonstrating stable nostril ROI alignment and reliable peak detection (red).}
    \label{fig:rr_tracker}
\end{figure}

We also estimate respiration rate using the FLIR Boson+ long-wave infrared (LWIR) camera, by capturing temperature differences of exhaled air around the nose and mouth. The pipeline tracks the face with the BlazePose model \cite{bazarevsky2020blazepose}, extracts a nostril region of interest (ROI), and applies a simple motion-stability check that rejects frames where head keypoints move beyond a small threshold or the pose confidence drops below a preset level. Within the ROI, temperature-filtered pixel intensities are temporally averaged over a short window; we then remove slow drift (e.g., sensor heating or slight head motion) by subtracting a moving-average baseline from the signal before smoothing and peak detection. Respiratory oscillations are recovered via peak detection with pose-aware thresholds and minimum peak spacing, and the system outputs instantaneous and averaged breathing-rate estimates along with a stability flag; a lightweight GUI as shown in Fig. \ref{fig:rr_tracker} visualizes the ROI, waveform, and detected peaks for operator verification. In practice, the method is most effective when exhaled air and facial skin are at least $\pm5\degree$\,F from ambient temperature, providing clear thermal contrast, and under these conditions we observe roughly 90\% agreement with a chest-worn ground-truth sensor in both indoor and outdoor tests.

\subsubsection{Event Camera}
We also developed an algorithm for respiration detection using event camera data. Using events allows us to leverage the high temporal resolution and low-light sensitivity of event cameras to detect chest motion and extract respiratory waveforms. When subjects are stationary and there is little background motion, we can successfully detect respiration rate within $\pm4$ bpm. However, this method remains highly sensitive to motion artifacts and often fails when subjects exhibit even minor movements, such as twitching, or when the chest is not clearly visible.

\section{TRAUMA DETECTION AND CLASSIFICATION}
\label{sec:vlm}

\begin{figure}
    \centering
    \includegraphics[width=\linewidth]{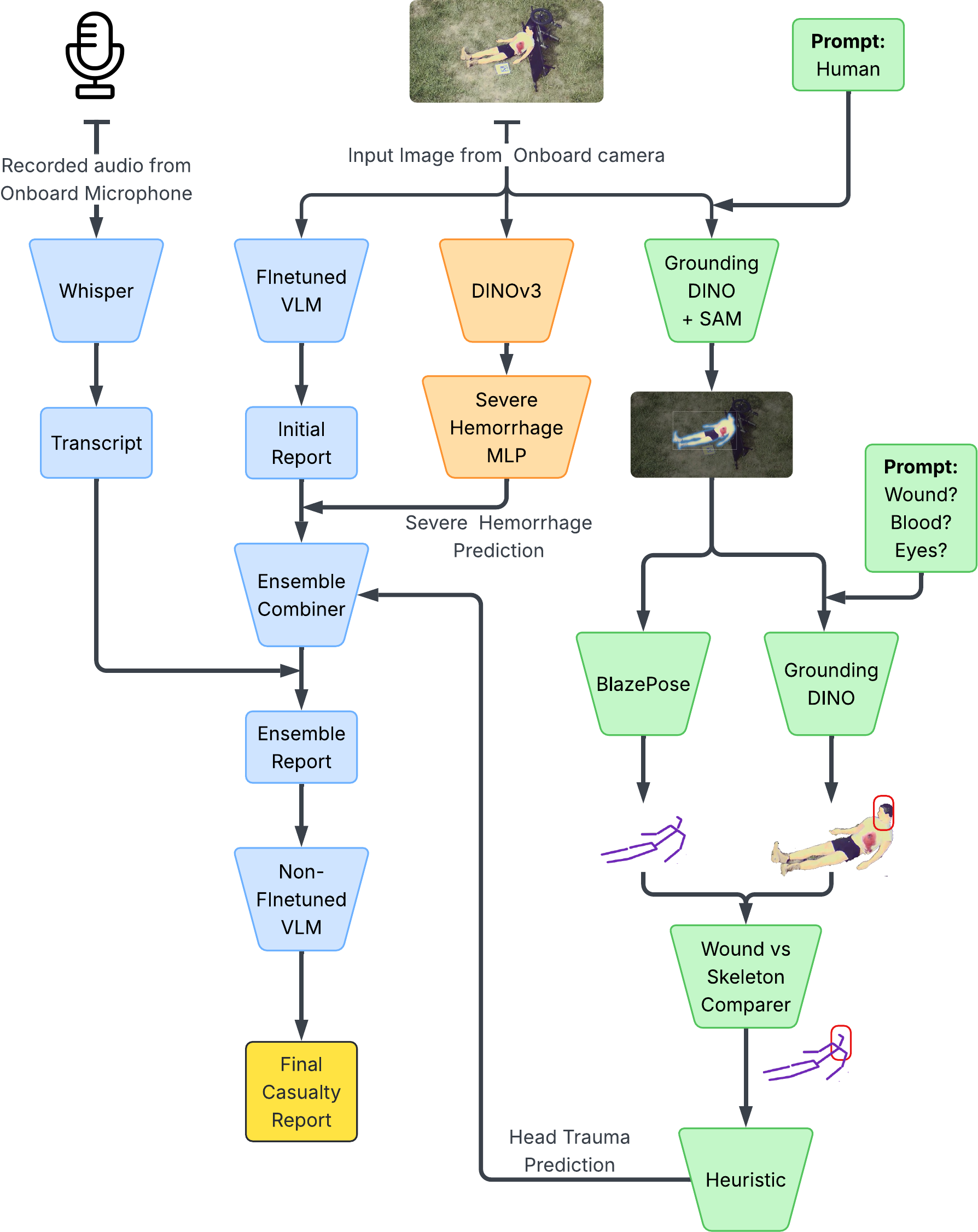}
    \caption{Overview of the triage assessment pipeline. Blue boxes indicate the VLM-based pipeline (NVILA for multi-view injury assessment and auxiliary reasoning), green boxes indicate the Grounding DINO-based injury localization pipeline, orange boxes indicate the DINOv3 classifier pipeline used for severe hemorrhage and the yellow box indicates the final casualty report generated by the triaging pipeline.}
    \label{fig:triage-pipeline}
\end{figure}

Another key component of our triage pipeline is the assessment of physical injuries. In this section, we describe how we use a combination of vision-language models (VLMs) and vision encoders to assess head, torso and extremity trauma, ocular alertness, severe hemorrhage, and respiratory distress from multi-view imagery and audio. Because large, labeled triage datasets are scarce, we build on pretrained foundation models that can be deployed fully on-premise hardware, without reliance on cloud connectivity.

\subsection{LLaVA-8B}



We first explored a VLM-only classification pipeline based on LLaVA-8B~\cite{liu2023visualinstructiontuning}. Vision-language models extend large language models to operate jointly on images and text, and have recently gained traction for medical and clinical applications~\cite{liu2025visual}. In general, these models are trained to maximize the likelihood of the next token given a sequence of embedded text and image tokens~\cite{clip}, often using transformer-style architectures~\cite{gpt4,brown2020language}.

We collected a dataset of roughly 50 unique victims with human-provided triage labels. For each victim, we capture two RGB views from different angles. We fine-tune LLaVA using Low-Rank Adaptation (LoRA) ~\cite{hu2022lora}, which inserts a small number of trainable low-rank matrices into the model. Only these additional parameters are updated during fine-tuning, preserving the base model’s capabilities while conditioning it on triage-specific visual cues.

The model is prompted to independently assess several trauma-related fields per image, including \emph{Head Trauma, Torso Trauma, Upper Extremity Trauma, Lower Extremity Trauma, Ocular Alertness, and Severe Hemorrhage}. LoRA fine-tuning significantly improves performance over the off-the-shelf model, but inference remains relatively heavy, requiring 24 GB of VRAM and around 30 seconds to process both views for a single victim.

\begin{table}[t!]
\centering
\caption{Per-field classification accuracy (\%) for three trauma-assessment methods: VLM, DINOv3 classifier, and the Grounding DINO pipeline on our collected validation dataset. Metrics are reported for severe hemorrhage, ocular alertness, and regional trauma (head, torso, upper and lower extremities); the best accuracy in each row is shown in \textbf{bold}.}
\label{tab:all_results}
\begin{tabular}{lcc>{\centering\arraybackslash}p{1.2cm}}
\toprule
\multicolumn{1}{c}{\textbf{Label}} &
\multicolumn{1}{c}{\makecell[c]{\textbf{VLM} \\\textbf{(\%)}}} &
\multicolumn{1}{c}{\makecell[c]{\textbf{DINOv3} \\\textbf{(\%)}}} &
\multicolumn{1}{c}{\makecell[c]{\textbf{Grounding DINO}\\\textbf{Pipeline (\%)}}} \\
\midrule
Severe Hemorrhage   & 47.76 & \textbf{50.5} & 43.9 \\
Alertness (Ocular)  & \textbf{80.60} & 56.9 & 56.8 \\
Trauma Head         & 79.10 & 69.1 & \textbf{83.5} \\
Trauma Torso        & \textbf{74.63} & 52.1 & 62.6 \\
Trauma Lower Ext.    & \textbf{64.18} & 35.6 & 48.2 \\
Trauma Upper Ext.    & \textbf{68.66} & 55.3 & 54.0 \\
\bottomrule
\end{tabular}
\end{table}

\subsection{NVILA-Lite-2B}

To enable full onboard deployment on the UGVs, we evaluated smaller VLMs and DINO-based classification models that fit within our compute budget. We tested several candidates, including Qwen2.5-VL~\cite{bai2025qwen2}, Phi-4 Multimodal~\cite{abouelenin2025phi}, and NVIDIA's NVILA-Lite-2B~\cite{liu2025nvila}, on our triage classification tasks. Based on these preliminary experiments, we selected NVILA-Lite-2B as our primary compact VLM due to its favorable accuracy-efficiency trade-off.

Unlike the larger LLaVA-8B model, NVILA-Lite-2B is small enough to support full-parameter fine-tuning. We cast each triage field as a visual question answering (VQA) problem: given one or more victim images and a question (e.g., \emph{``is there trauma to the torso?"}), the model must answer \emph{``yes"} or \emph{``no"}. During training, we mix our triage dataset with a standard VQA corpus~\cite{antol2015vqa} to avoid catastrophic forgetting and preserve the model's general reasoning ability.

At inference, we ask a series of short, field-specific yes/no questions over two views per victim. This framing keeps the context length small and reduces prompt complexity, which we found to be important for stable performance of compact VLMs. NVILA-Lite-2B is also used for several auxiliary reasoning tasks that refine the overall triage assessment:
\begin{itemize}
    \item \emph{Speech assessment:} audio from the UGV microphone is transcribed using Whisper~\cite{radford2022robustspeechrecognitionlargescale}, and the transcript is passed to NVILA to check whether the victim's speech is coherent and whether they verbally indicate any injuries.
    \item \emph{Manikin detection:} because the challenge includes both manikin and live actors, we use NVILA to classify whether the subject is a manikin, which informs how we interpret vitals (e.g., heart rate should be zero for non-animated manikins).
    \item \emph{Natural-language descriptions:} a non-fine-tuned NVILA checkpoint generates a concise description of the victim and surroundings, which helps first responders locate specific victims using contextual cues. 
\end{itemize}

\subsection{Grounding DINO Pipeline}

We also explored a more traditional vision pipeline built on DINOv2~\cite{oquab2023dinov2} using Grounding DINO~\cite{liu2024grounding}, an object detector grounded in natural language. Our goal is to localize injuries on specific body regions and convert them into structured trauma fields.

Given an input image, we first prompt Grounding DINO to detect the human in the scene. The resulting bounding box is passed to SAM2~\cite{ravi2024sam} to obtain a segmentation mask that removes most environmental clutter. The masked victim image is then processed by BlazePose~\cite{bazarevsky2020blazepose} to estimate a full-body skeleton and joint locations in pixel coordinates.
We feed the same masked image back into Grounding DINO with injury-related prompts such as \emph{``wound", ``blood"} and \emph{``amputation"}. For each detected injury, we take the center of its bounding box and compare it against body-part groups derived from the BlazePose skeleton (e.g., head, torso, upper extremities, lower extremities). If an injury center lies within a distance threshold of a given body region, we mark that region as having the corresponding trauma.

The pipeline yields interpretable, spatially grounded predictions for trauma fields and is particularly effective for head trauma, where localized injuries tend to be visually salient. Its performance relative to the VLM-based approaches is summarized in Tab.~\ref{tab:all_results}.

\subsection{DINOv3 Classifier}

Despite reasonable performance from both the VLM and the Grounding DINO pipeline, severe hemorrhage remained challenging to classify reliably, as shown in Tab.~\ref{tab:all_results}. To better target this field, we trained a dedicated image classifier on top of the DINOv3~\cite{simeoni2025dinov3} vision encoder.

We use DINOv3 to extract high-level image features from each victim view and attach separate lightweight multi-layer perceptron (MLP) heads for binary classification (presence or absence) on each of the classes shown in Tab.~\ref{tab:all_results}. The model is trained on our curated dataset of labeled triage images, using standard data augmentation to improve robustness to viewpoint, lighting and background variations.

\subsection{Datasets}

In order to train these models, we used a combination of datasets provided by DARPA for the challenge and data we collected ourselves. For our in-house datasets, we applied moulage to dress manikins and volunteer actors as injured victims. Our initial dataset, used to LoRA-fine-tune LLaVA-8B, contained around 50 distinct victims with a variety of injury patterns. In the second year of the competition, we expanded our data collection to roughly 500 distinct victims. This larger dataset was used to fine-tune NVILA-Lite-2B and the DINOv3 classifier heads.


\section{DISCUSSIONS AND LESSONS LEARNED}
In sections \ref{sec:va} and \ref{sec:vlm} we discussed the methods we developed for vitals assessment and physical injury detection as part of a complete primary triage assessment. In this section, we discuss what was actually deployed in the first and second years of the DARPA Triage Challenge. Before deploying our system in the competition we set up our own triage scenarios using moulage to simulate real injuries and having volunteers modulate their breathing in different ways and drew our conclusions from those scenarios. 



\subsection{Vitals Assessment}

To asses the accuracy of the heart rate and respiration rate detection methods, we used ground truth sensors, a heart rate and respiration rate belt respectively, worn around the chest of the volunteer. We tested the methods while the volunteer simulated realistic triage scenarios, including moving, occluding their chest with their arms and lying, sitting or standing in different orientations with respect to the sensor. For each method, we considered both accuracy and robustness to these conditions when deciding whether to deploy it in the competition system.

For heart rate detection, we found that the rPPG method was not suitable for deployment in our triage system. In addition to a long runtime (often about 45 seconds per inference), the method frequently failed when the UGV did not obtain a clear view of skin, leading to unreliable heart rate signals in realistic field conditions.
The MTTS-CAN network falls victim to the same issue. The training data for MTTS-CAN is mostly made up of videos where there is a clear view of the face which may not be true in real world triage scenarios. 
Although MTTS-CAN often produced signals that appeared reasonable when compared to ground truth and was initially deployed in the year-one system, further analysis suggestsed that it tended to predict a plausible average heart rate in the 72-78\,bpm range regardless of the true value.
Finally, we tried a mmWave radar which gets promising results when the person is sitting still, but when the person is not sitting still the heart rate signal can get drowned out from the movement. We nevertheless deployed this as the heart rate detection in year two of the competition because of the constraint of operating at night, MTTS-CAN requires 10 seconds of imagery which is not possible in low light conditions. To mitigate the motion problem, we used the speaker onboard the UGV to ask the victims to remain still before taking the measurements with the radar.

For respiration rate, we initially tested a pulsed coherent radar. This works well when there is a clear view of the front of the chest, views from the side do not give enough of a signal to deduce respiration rate. We deployed this in the first year system since it is accurate when the UGV can view the chest of the person. For the MTTS-CAN respiration rate prediction, we found similar issues as with heart rate prediction. It seems to perform well but is consistently predicting an average range of respiration rates for people. For the first year system, we felt this augmented the shortcomings of the PCR and deployed MTTS-CAN alongside it. We report the MTTS-CAN result when the PCR method could not produce a result. In year two of the competition we developed the LWIR method for respiration rate detection. Since this method relies on tracking facial features it is robust to the victims motion but tracking these features is difficult if there is not a clear view of the face. This method will also fail when the ambient temperature is near that of the human body since there will not be a distinct enough difference between the background temperature and the temperature of the victim's breath to get a signal. Since this method is quite robust, we did deploy it in the year two system. Finally, we also deployed the mmWave radar to fill in the shortcomings of the LWIR method, but this method has the same downfalls as using it for heart rate detection. Thus, in the year two system we report the mmWave prediction for respiration rate only when the LWIR method cannot produce a result.

\subsection{Injury Detection and Classification}
For the year one system we deployed the LoRA-fine-tuned checkpoint of the LLaVA model as our injury detections and classification part of the system. At the time, the smallest available checkpoint, LLaVA-8B used around 24 GB of VRAM, meaning it could not be deployed on the UGV which only has 20 GB of VRAM. Instead, we would send select images, via our distributed database, from the UGV to a basestation with two NVIDIA RTX 4090 GPUs where our LLaVA model was running. The UGV would also send the vitals assessments along with the images. The VLM would assess head, torso and upper and lower extremity trauma as well as severe hemorrhage, respiratory distress and ocular, verbal and motor alertness. Once this was completed, the victim assessment or scorecard could be sent to first responders or to DARPA to be scored. The major shortcoming of this method is that it is centralized, and the basestation had to do the triage inference of multiple robots making it a bottleneck, especially since processing could take up to 30 seconds per victim.

The goal for the year two system was to push all the triage inference onboard the UGVs. We expanded beyond just VLMs and tested other models that would allow for this. As a result, in the second iteration of the system we were able to deploy the NVILA-Lite-2B VLM, a DINOv3 classifier and Grounding DINO simultaneously onboard the UGV. We benchmarked the three models against each other as shown in Tab.~\ref{tab:all_results}. We use each model where it performs best, meaning the VLM was used for ocular alertness, torso trauma, lower extremity, and upper extremity trauma, the DINOv3 classifier was used for severe hemorrhage prediction and the Grounding DINO and BlazePose pipeline was used for head trauma prediction. Finally, we used the VLM to classify the text from the Whisper model as coherent to determine verbal alertness and track the victim's joints with BlazePose over three frames to determine motor alertness. The combination of these methods is depicted in Fig.~\ref{fig:triage-pipeline}.

\subsection{Lessons Learned}

To better triage with the VLM, we made some improvements at runtime to help the model classify better. From the first year's system we observed that long VLM context length degrades model performance on the available comparatively small open-source models. Thus, we shifted from trying to give as much context in the prompt as possible to a series of yes or no questions, such as \emph{``is there a head injury”, ``is there blood”}, etc. If at least one of the queried images respond with \emph{``yes”} for a specific trauma type, the returned report denotes \emph{``presence”} for said trauma, which also greatly reduced the inference time. 

We also learned that the views taken of the victim are crucial for accurate triage. To this end, in year two we used lenses with a larger focal length to place more pixels on the victim and improve spatial resolution. We also learned that victim motion is inevitable and if the respiration and heart rate monitoring algorithms cannot account for this, the accuracy will not be great in real-world deployments. This is why we developed the LWIR based respiration rate algorithm to be robust to this motion.

\section{CONCLUSIONS}
In this technical report we discussed the iterations of our air-ground multi-robot team for rapid remote triage as part of the DARPA Triage Challenge. We developed and deployed methods for non-contact heart rate and respiration rate detection and combined them with vision-language models and large vision encoders to locate and classify victims' physical injuries. The final version of our system was able to do all of this in real time onboard the UGVs themselves. 

To improve the system in the future we need to improve the performance of the learned models which means more data to train with. To gather our own data for finetuning our models, we had to perform our own data collection. Data collection by itself is incredibly time intensive. As a result, a current research direction being explored is leveraging generative models like Stable Diffusion~\cite{rombach2022high} to generate images that extend our sparse data distribution in a way that is useful for training more accurate downstream image classifiers. 

We are also working toward tighter air-ground cooperative autonomy so that little to no human intervention is required. In particular, we aim to use UAV imagery for traversability analysis and global planning context, enabling UGVs to navigate autonomously to victims while maintaining the triage performance demonstrated here.






\bibliographystyle{IEEEtran}

\bibliography{bibliography}
\end{document}